# Machine-Learning-Enhanced Soft Robotic System Inspired by Rectal Functions for Investigating Fecal incontinence

Zebing Mao[1*], Sota Suzuki[2], Hiroyuki Nabae[2], Shoko Miyagawa[3], Koichi Suzumori[2,4], Shingo Maeda[2,4]

[1]Faculty of Engineering, Yamaguchi University, Yamaguchi, Japan
[2]School of Engineering, Tokyo Institute of Technology, Tokyo, Japan
[3]Faculty of Nursing and Medical Care, Keio University, Kanagawa, Japan
[4]Research Center for Autonomous Systems Materialogy (ASMat), Institute of Innovative Research, Tokyo Institute of Technology, 4259 Nagatsuta-cho, Midori-ku, Yokohama, Kanagawa, 226-8501, Japan

Corresponding: mao.z.aa@yamguchi-u.ac.jp

## Abstract

Fecal incontinence, arising from a myriad of pathogenic mechanisms, has attracted considerable global attention. Despite its significance, the replication of the defecatory system for studying fecal incontinence mechanisms remains limited largely due to social stigma and taboos. Inspired by the rectum's functionalities, we have developed a soft robotic system, encompassing a power supply, pressure sensing, data acquisition systems, a flushing mechanism, a stage, and a rectal module. The innovative soft rectal module includes actuators inspired by sphincter muscles, both soft and rigid covers, and soft rectum mold. The rectal mold, fabricated from materials that closely mimic human rectal tissue, is produced using the mold replication fabrication method. Both the soft and rigid components of the mold are realized through the application of 3D-printing technology. The sphincter muscles-inspired actuators featuring double-layer pouch structures are modeled and optimized based on multilayer perceptron methods aiming to obtain high contractions ratios (100 %), high generated pressure (9.8 kPa), and small recovery time (3 s). Upon assembly, this defecation robot is capable of smoothly expelling liquid faeces, performing controlled solid fecal cutting, and defecating extremely solid long faeces, thus closely replicating the human rectum and anal canal's functions. This defecation robot has the potential to assist humans in understanding the complex defecation system and contribute to the development of well-being devices related to defecation.

## 1. Introduction

Fecal incontinence (FI) is characterized by the involuntary discharge of bowel contents, including gas, mucus, and both liquid and solid feces, indicating an inability to control defecation [1]. The etiology of fecal incontinence encompasses a diverse array of pathogenic mechanisms, commonly attributable to rectal anomalies, dysfunction of the sphincter muscles, nerve impairment, and various other medical or physiological conditions [2]. In response to these challenges, the advent of physiological organ simulators and computational models marks a significant stride towards offering alternative approaches for expediting the evaluation of potential treatment methodologies during preliminary development phases and enhancing the comprehension of biomechanical processes underlying diverse disease states [3][4]. Computational models, while invaluable in the study of human physiology and the simulation of organ functions, often fall short in capturing the complex interplay among multiple organs. This limitation stems primarily from the inherent challenge of accurately replicating the intricate, dynamic interactions that occur within the human body. Such interactions involve not only the mechanical and physical forces but also the biochemical processes that govern organ function and interorgan communication. Consequently, the lack of realism in these models can hinder our ability to fully understand the holistic functioning of the body's systems and to predict the outcomes of medical interventions with high fidelity [5].

Soft robotics have recently emerged as a cutting-edge approach for simulating or enhancing the functions of human organs [6][7][8][9][10]. For instance, implantable soft robotic devices have demonstrated promising outcomes in augmenting cardiac function in cases of isolated heart failure, whether affecting the left or right side of the heart [11]. Similarly, stomach robots have been developed to provide a controlled testing environment for innovative food products and pharmaceuticals [12][13], while the biomimetic robotic soft esophagus (RoSE) serves as an in vitro model for evaluating endoprosthetic stents designed to manage dysphagia [14]. These developments underline the potential of soft robotics as both a complementary and supportive technique in conjunction with mathematical models, particularly for exploring human physiology and validating medical interventions [15]. Notably, anal sphincter (AAS) devices utilizing a circular cuff design have been employed in the treatment of fecal incontinence in humans [16][17]. Researchers utilized different actuation techniques: rigid clamping mechanism [18], fluidic actuators [19][20][21], cable-driven actuator [22], magnetic actuator [23], shape memory alloys (SMA) actuator [24] etc., to investigate their ability to reproduce sphincter pressures observed in clinic. Also, other actuators have shown the protentional for occluding the canal [25][26][27][28]. Moreover, other researchers developed a reusable phantom model that combines realistic anatomy and realistic tissue properties in a cost-effective manner [29]. Although present endeavors have primarily revolved around the AAS device, research focused on replicating the defecatory system and delving into the mechanisms associated with fecal incontinence (FI) is notably limited. The scarcity of studies in this area can primarily be ascribed to the stigma and social taboos related to the subject, which significantly impedes progress in understanding and addressing FI through innovative soft robotic simulations.

In this study, we propose an entirely soft and lightweight pneumatically-driven defecation robot, including several soft ring-shaped actuators within soft cases, to replicate rhythmic peristaltic motion and opening-occlusion motion of the anal at the symmetrical level. Fig. 1 shows a detailed visual analysis of the rectal defecation process, alongside a conceptualized soft robotic system designed to replicate these physiological movements. The muscular anatomy includes the puborectalis muscle, which plays a pivotal role in continence, as well as the internal and external sphincter muscles, crucial for the regulation of stool passage (Fig. 1**A**). In our study, we set the anorectal angle (ARA) to 165 degrees, which is helpful for fecal defecation. The cross-sectional view of the rectal architecture highlights its composite structure, which consists of the mucosal surface, the underlying intestinal gland, and the muscular layers, specifically the circular and longitudinal muscles (Fig. 1**B**). The rectum is a muscular tube consisting of a continuous longitudinal muscle layer that connects with the underlying circular muscle. The longitudinal and circular muscle are to propel fecal matter along the gastrointestinal tract and contracts radially to crush and grind the faeces [30]. Three distinct defecation scenarios sometimes occur for human kinds: (i) Diarrhea, illustrating the process of expelling liquid faeces, (ii) Cut faeces, showcasing the expulsion of solid faeces that are abruptly truncated, and (iii) Defecate long faeces, demonstrating the continuous expulsion of a solid stool (Fig. 1**C**).In this study, we simulate these scenarios to understand the rectum's role in defecation. The peristaltic progression of fecal matter from the sigmoid colon into the rectum is depicted through a dynamic cross-sectional visualization of the fecal bolus, showing a decrease in cross-sectional area as indicated by peristaltic contraction rate ($\alpha$) (Fig. 1**D**). The successive cross-sectional illustrations (1~5) visually quantify this reduction from 100% to approximately 30%. Additionally, a segmented soft pouch actuator system is designed to mimic the natural peristaltic contractions of human bowel movements, with the actuator segments demonstrating varying degrees of contraction to replicate the rectum's complex muscular movements during defecation.

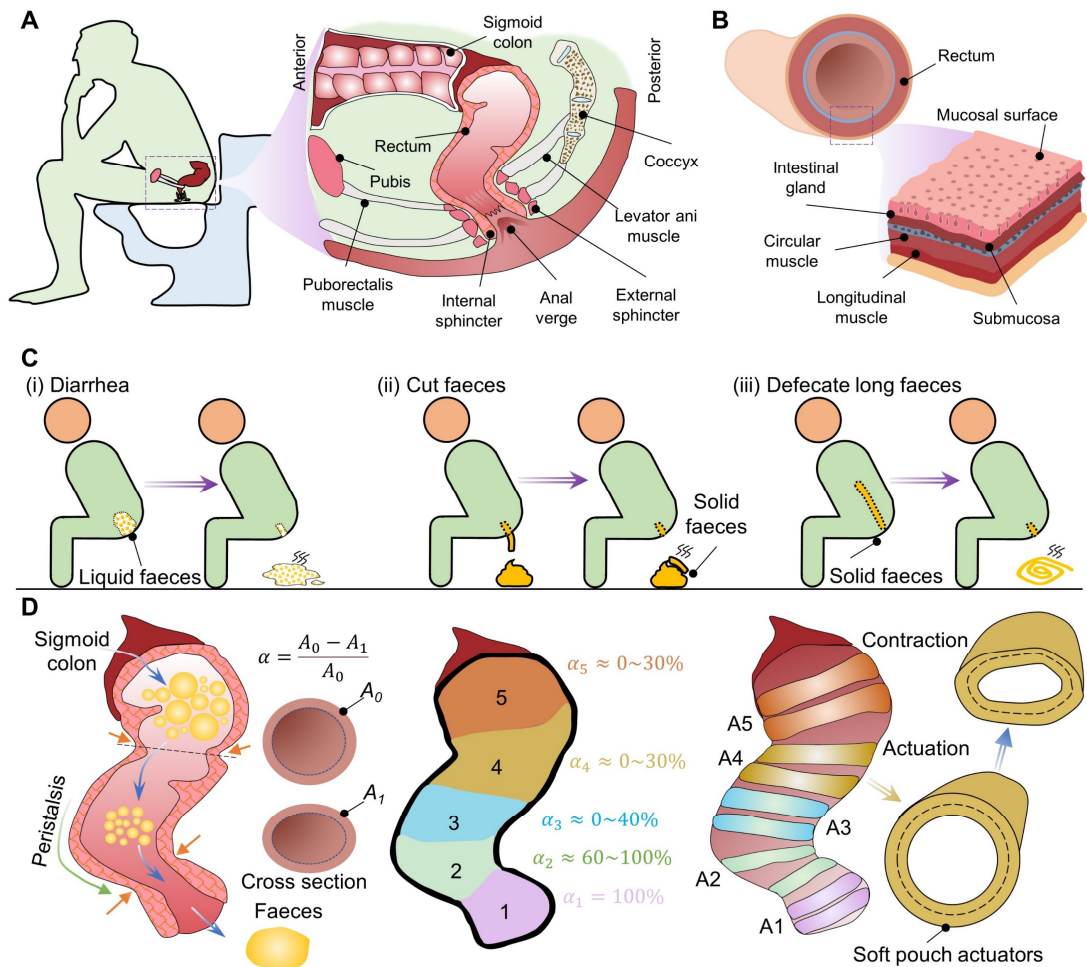

**Fig. 1** Conceptualization of a Soft Robotic System for Simulating Defecation. **A:** Schematic overview of key physiological components of the defecation system. **B:** Anatomical structure of the muscle layer of the rectum. **C:** Three unique defecation scenarios encountered in humans. **D:** Defecation process within the rectum, including the contraction rate of each rectal section and the strategic placement of actuators within the robotic model ($A_0$ represents the initial area when the actuator is fully extended; $A_1$ denotes the contracted area during actuation).

## 2. Results and discussion

### 2.1 Rectum model

The tensile properties of human rectal tissue demonstrate viscoelasticity, as evidenced by increased material stiffness at larger displacements [31][32]. To develop materials with elastic moduli comparable to the human rectum, we investigated a range of soft and hard materials, including AR-M2, Elastic 50A, AR-G1L, Agilus30-1, and Ecoflex 00-30 (Fig. 2**A & B**). AR-M2 was classified as a hard material, while the remaining materials were considered soft. We evaluated the soft materials by analyzing their stress-strain relationships, preparing samples according to the dumbbell-type specimen standards of JIS No. 7. For the construction of the rectum model, we employed various methods, including 3D printing and mold replication (FigS.1). The result indicates that Ecoflex 00-30 most closely replicates the human rectum's stress and strain characteristics. The fabrication process required multiple molds, including the inner, bottom, upper, and cross molds (Fig. 2**C**). The inner mold was produced in two halves and was subsequently joined with adhesive due to its length. These molds were assembled into a single unit, which the prepared Ecoflex 00-30 mixture (A: B = 1:1) was poured into. This assembly was then cured at 60°C for one hour before being demolded (Fig. 2**D**). To enhance the installation and testing process of subsequent actuators, we partitioned the model into five segments (denoted as S-1 through S-5). The radii at the entrance on both sides of S-2 were set to 17 mm and 21 mm, respectively. Similarly, for S-3, the entrance radii were designed to be 21 mm and 28 mm. In the case of S-4, the entrance radii on both sides are 28 mm and 32 mm, respectively. The radius of S-5, which is cylindrical in shape, is 32.5 mm. The constructed rectum model is segmented into five parts, representing the short anal canal, the upper, and lower rectum, which exhibit complex curvatures and twisted angles along the axial direction of the rectum (Fig. 2**E**). These five segments were also replicated utilizing the described methods (Fig. 2**F**).

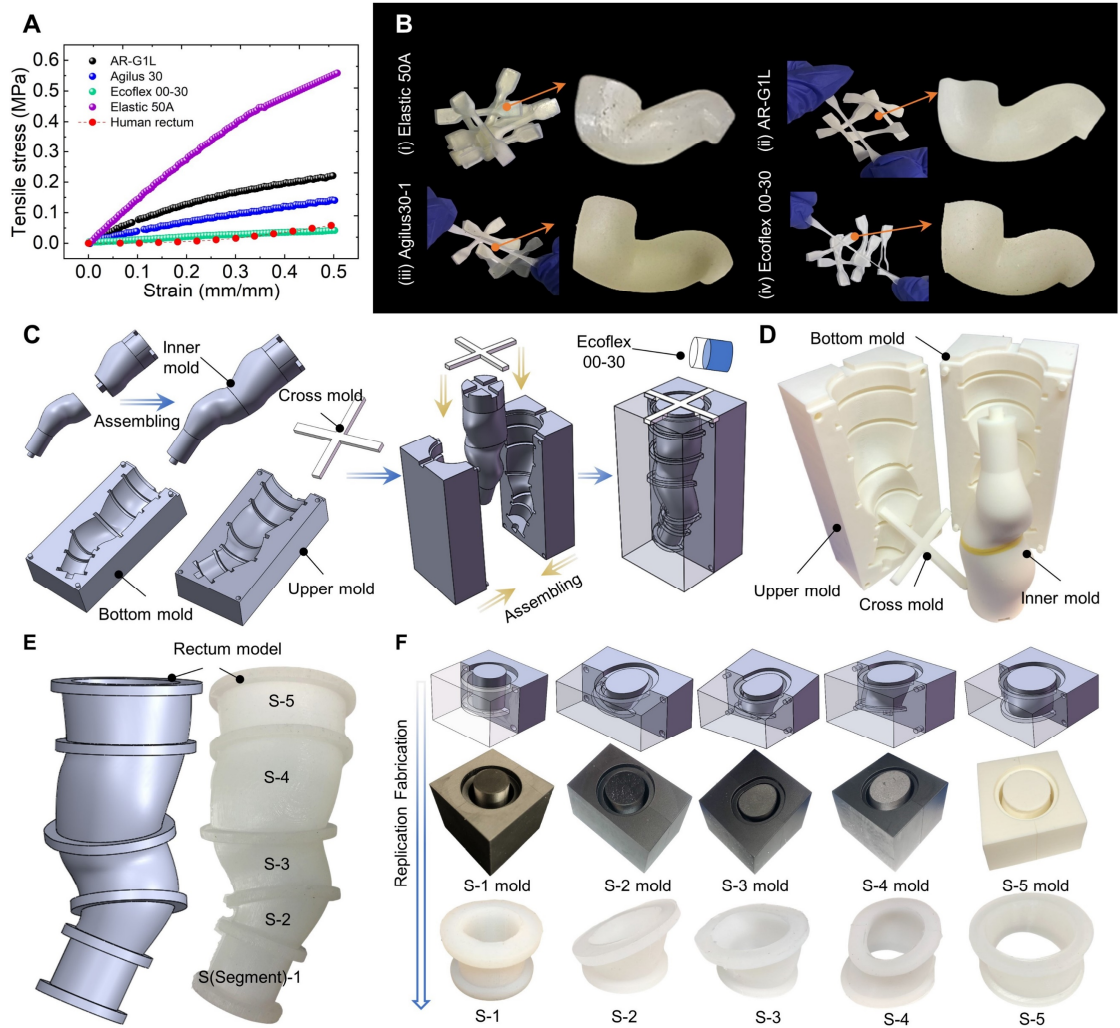

Fig. 2 Material Selection and Fabrication Method of the rectum model. **A:** Stress-strain characteristics of various materials compared with the human rectum. **B:** Optical images of rectum models printed from four types of soft elastic materials. **C:** Fabrication procedures for the rectum model. **D:** Optical image of the rectal molds. **E**: Photos of rectum model alongside a 3D model. **F:** Fabrication method and optical images of each segment.

## 2.2 AAS muscles-inspired actuators

Inspired by the motion of the anus actuated by the anal sphincter muscles (AAS), we have designed ring-shaped actuators to simulate the external sphincter's function in opening and occluding the anus (Fig. 3**A**). The external anal sphincter, a voluntary muscle under conscious control, enables individuals to contract and relax it at will, facilitating control over bowel movement timing. To mimic this functionality, we introduced a soft pneumatic pouch actuator based on the principle that introducing air into a flat pouch causes it to contract longitudinally, as the air prompts lateral expansion of the actuator [33][34]. Aiming for durability, we selected three types of thermoplastic materials: polypropylene (PP), BOPP (Biaxially Oriented Polypropylene Film), and HDPE (High Density Polyethylene), each measured at 130 mm in length and 25 mm in width (Fig. 3**B**). To ensure a reliable connection to the pneumatic pumps, we employed a commercial adaptor (VRF206, MonotaRO, Japan), coupled with a relatively rigid polyurethane tube (PISCO, inner diameter: 2 mm, outer diameter: 4 mm) and reinforced by a soft silicone tube (inner diameter: 3 mm). Subsequently, we applied a cyanoacrylate adhesive (Loctite 401) to seal the clearance around the connecting parts. In the experimental setup, we applied a control signal characterized by a square wave, with a period of 4 seconds and a duty cycle of 50%. To prevent potential damage from high pressure, a directional valve was employed to vent the system, safeguarding the soft actuator from overexpansion. Fig. 3**C** illustrates the durability of three developed pouch actuators under applied pressures ranging from 20 kPa to 100 kPa. The results indicate that pouch actuators made from polypropylene (PP) exhibit superior durability, outperforming those made from BOPP (Biaxially Oriented Polypropylene Film) and HDPE (High Density Polyethylene). Specifically, PP-made actuators maintain functionality for over 5400 cycles at pressures below 40 kPa, while their lifetime decreases significantly at pressures exceeding 60 kPa. In contrast, actuators made from BOPP and HDPE fail at pressures exceeding 50 kPa and 30 kPa, respectively. To convert the linear displacement of pouch actuators into circumferential contractions, we construct ring-shaped actuators by rolling the flat pouch actuators and sealing the overlapped areas with double-coated adhesive tape (No.5000NS, Nitto) (Fig. 3**D&E**).

Firstly, we constructed three variants of actuators to assess the influence of a soft shell on their performance: Type I (a pure pouch actuator), Type II (encased in PET (polyethylene terephthalate) paper), and Type III (encased in an A80 ring). The fabrication of the designed PET paper ring and A80 ring was performed using a laser cutting machine (TROTEC Speedy 100) and a 3D printer (Form 3+, Formlabs), respectively. To evaluate the pressure output by these actuators, we devised a simple pneumatic circuit comprising a power module, a pressure sensor, and a rectal balloon catheter (46Fr 75mL, Create Medic Co.,Ltd) (Fig. 3**F**). The testing procedure entailed inflating the catheter within the testing apparatus to achieve a pressure of 10 kPa. Upon closing the valve and resetting the pressure sensor to 0, the actuators were mounted on the catheter, allowing for the evaluation of their performance by inflating them to a specified pressure. Fig. 3**G** illustrates the three types of ring-shaped actuators, operating in both static and inflated states (input pressure: 30 kPa). We determined the values of $\alpha$ for each type of actuator under varying applied pressures (Fig. 3**H**). The results indicate that $\alpha$ increases with the application of higher pressures, reaching peak values beyond 20 kPa. Specifically, the maximum contraction ratio for Type I actuators approaches 100%. Conversely, the maximum contraction ratios for Types II and III are significantly lower, attributable to the stiffness differences between the PET paper and the A80 ring. Additionally, we assessed the pressure generated within the anal canal, revealing a direct correlation between increased actuator pressure and sensor readings (Fig. 3**I**). It is noted that the type III ranks the first place, in which the relatively rigid cover can restrict the spherical actuation and thereby causing much more pressure to the sensor. The peak pressure recorded for Type III actuators reached 6.2 kPa at an applied pressure of 30 kPa. Regarding recovery time, Type III actuators demonstrated an enhanced ability to revert to their original shape, attributed to the rigidity of their materials (Fig. 3**J**).

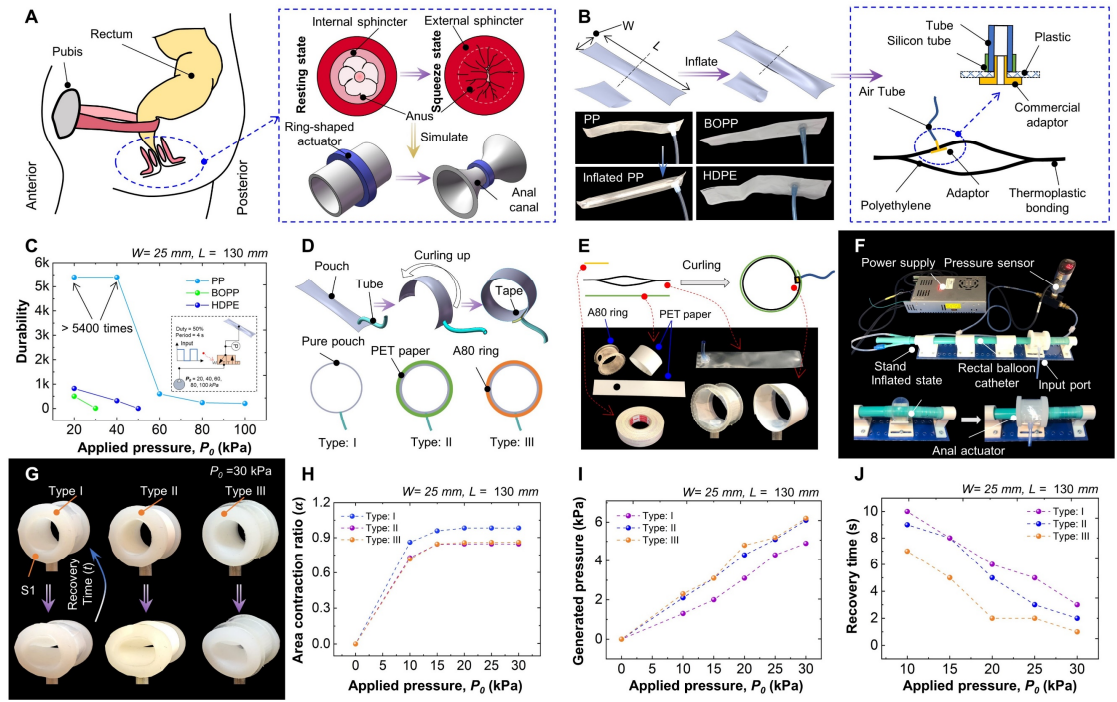

Fig. 3 Design, fabrication, and performance characteristics of AAS muscles-inspired actuators. **A:** Mimicking process of AAS muscle actions using ring-shaped actuators. **B:** Pouch actuators made of three different materials and their connectors to outer. **C:** Durability assessment of the three pouch actuators in response to varying levels of inflation pressure. **D:** Fabrication procedure of ring-shaped actuators with various encasements. **E:** Photos of the pouch actuator encapsulated by different materials. **F:** Photos of measuring system for the actuators' generated pressure. **G:** Comparative photos of AAS muscles-inspired actuators in both static and inflated states for types I, II, and III. **H:** Performance comparison of three types of actuators in terms of area contraction ratio. **I:** Performance comparison of three types of actuators in terms of generated pressure. **J:** Performance comparison of three types of actuators in terms of recovery time.

## 2.3 Machine learning-enhanced modeling and optimizing process

Aim to identify actuators exhibiting optimal performance characterized by significant contraction ratios, high generated pressures, and minimal recovery times, we engineered a variety of actuators with diverse structures, cover types, lengths, widths, and turns under the static pressure ranging from 0 ~30 kPa. Subsequently, we conducted experiments to assess their performance based on three criteria: recovery time ($t$), contraction ratio ($\alpha$) and generated Pressure ($P_g$). Fig. 4**A** shows the workflow for the modeling and optimizing process. The experimental setup comprised an air compressor, pressure gauge, pneumatic actuators, a timer, and a catheter. Our data analysis protocol included data gathering, computation, model training, prediction, and parameter refinement, aiming to derive the most effective input parameters for AAS-inspired pouch actuators. Fig. 4**B** shows various actuator types with distinct structural properties, ranging from Type 1 to Type 4, which can be fabricated by the thermosetting process (Video S1&FigS.2). We also explored the impact of length on the contraction ratios of pouch actuators under diverse pressure scenarios (Fig. 4**C**). In Fig. 4**D,** the actuators with varying widths (15 mm ~ 30 mm) were studied.

Additionally, the effect of varying the number of turns (1~3) on the performance of Type 1 pouch actuators was systematically assessed (Fig. 4**E**& Video S2).

Considering to modeling process, preliminary steps involved data preprocessing to accommodate both categorical and numerical data types, followed by the segmentation of the dataset into training and testing subsets. Here, we adopted a Multilayer Perceptron (MLP) model to estimate the output of the acutators since they have been used in several fields [35][36]. The model incorporating two hidden layers with 150 and 100 neurons, respectively was implemented. We used the MAE (Mean Absolute Error), MSE (Mean Squared Error), $R^2$ (coefficient of determination) to evaluate the accuracy of the estimated models, which are 0.41, 0.59, and 0.99 respectively. Furthermore, we examined the error distribution for contraction ratio, generated pressure, and recovery time of the pouch actuators (Fig. 4**F&G&H**). The error distribution histograms reveal that the majority of errors cluster near a value of zero, suggesting that our model achieves a high level of prediction accuracy. The investigation of feature importance (Fig. 4**I)** highlights that cover material, coil turns, and structural design significantly affect the pouch actuators' output parameters. Subsequently, this highly accurate model was employed to identify optimal inputs within the bounds of physical constraints. To articulate this objective, we utilized a straightforward function, which is described as follows:

$$f = Min[t/(P_g * \alpha)] \tag{1}$$

By the application of the modeling and optimization process, we finally determined the optimal values for six input features: a length of 130 mm, a width of 25 mm, a single coil turn, an input pressure of 30 kPa, a paper material for the cover, and a Type 4 structural design. The comparison with other actuators (Table S.1) reveals that our actuators can achieve complete occlusion in large-diameter hollow tubes, typically those with diameters greater than 30 millimeters but less than 50 millimeters.

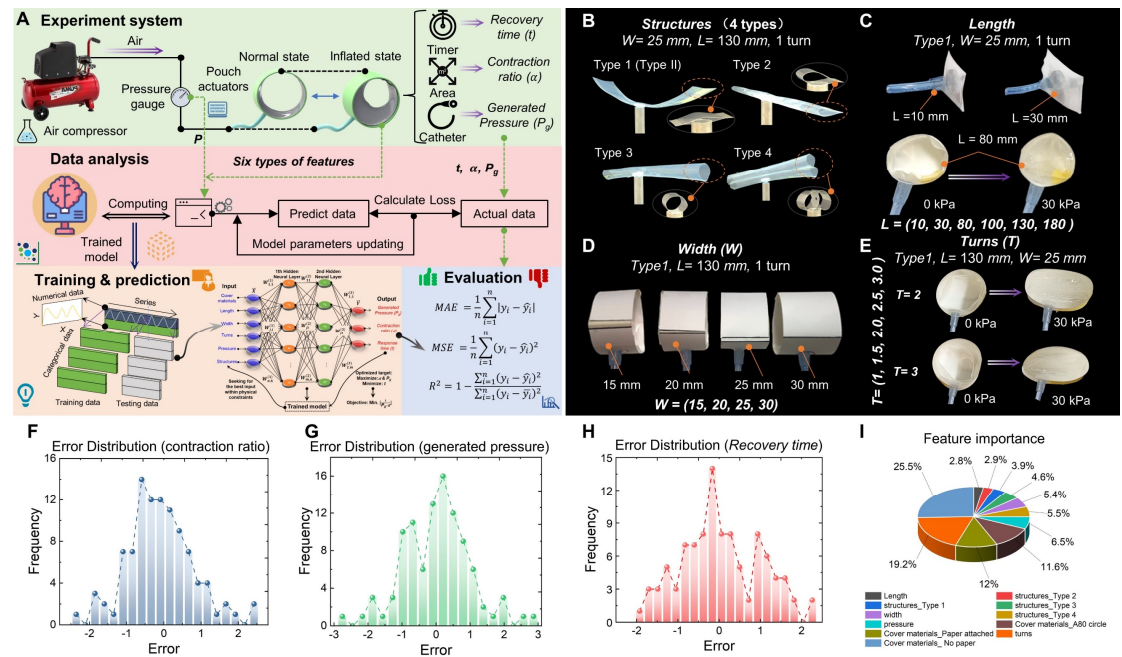

Fig. 4 Machine learning-enhanced optimization process characterized by a series of systematic steps and outcomes. **A:** Workflow of MLP (Multilayer Perceptron)-based learning, encompassing the experimental system setup, data analysis, model training, predictive assessment, and performance evaluation. **B:**

Feature 1: pouch actuators with shapes of different cross-sections. **C:** Feature 2: Length *(L)*; **D:** Feature 3: width (*W*); **E:** Feature 5: turns (*T*)**; F**: Error Distribution for contraction ratio; **G**: Error Distribution for generated pressure; **H:** Error Distribution for recovery time; **I:** Analysis of feature importance.

## 2.4 Rectal module

The rectal module comprises a rectum model, five double-layer actuators, and five casings both rigid and soft (Fig. 5). Fig. 5A illustrates the components and the assembly process of the rectal module. For segment S-1, we selected the ring-shaped actuator coupled with a paper ring due to its superior occlusion capabilities, characterized by a higher area contraction ratio. For segments S-2 through S-5, the A80 ring was chosen, as 3D printing technology enables the production of soft shells in various shapes. The actuators, denoted as A1 through A5, are highly adaptive, double-layer, pouch-based designs. They are engineered to replicate a simplified system of anal opening-occlusion motion (A1), mimic the amplitude and width variations of the smooth muscles in the rectum to facilitate peristaltic motion (A2 to A4), and prevent fecal backflow (A5), respectively. To control radial expansion, four soft cases (SC1 to SC4) and one rigid case (RC5) were fabricated. Each segment of the rectum model incorporates curled pouch actuators and a 3D-printed soft shell, enabling radial contraction. The assembly of the soft actuators and covers was achieved using needles and threads.

Fig. 5**B** shows the assembled rectal module from top and elevation views. The prepared actuators and cases are shown in Fig. 5**C.** Fig. 5**D** reveals the actuators in their initial and inflated states, demonstrating that actuators A1, A2, and A3 can tightly occlude segments S-1, S-2, and S-3, respectively, while A4 and A5 do not achieve full occlusion of S-4 and S-5 (Video S3). This performance aligns with the expected range illustrated in **Fig. 1**. To assess the actuator system's effectiveness in expelling simulated fecal matter, we employed ordinary clay to mimic faeces, shaping it into elongated forms and affixing it to the device. Upon application of pressure, it was observed that the actuator system could segment the simulated faeces, facilitating the automatic detachment of the cut portion (Fig. 5**E &** Video S4). Further analysis of the pressures generated by the system revealed that they are comparable to the minimum residual pressure and maximum squeeze pressure observed in the female anal region, as indicated in Fig. 5**F** [37]. By adjusting the pressure within a range of 30 to 120 kPa, we matched the pressure levels found in the human rectum (female). Regarding contraction ratios, it was noted that segments S-2 and S-3 could effectively close, as viewed from both perspectives A and B. The contraction ratios for S-4 were measured at 0.98 and 0.96 in views A and B, respectively (Fig. 5**G**). S-5, designed to prevent backward movement of the simulated stool and secure the entire model, maintained a certain degree of contraction, with ratios around 0.64 and 0.66. While S-2 and S-3 achieved maximum contraction ratios of 1, S-4 also exhibited satisfactory contraction. Because of its radius of 32.5 mm, S-5 displayed the lowest contraction ratio, aligning with its primary function of simulating fecal retention rather than complete closure, serving mainly as a barrier. Lastly, the recovery time of each segment following decompression (at an applied pressure of 30 kPa) was examined (Fig. 5**H**).

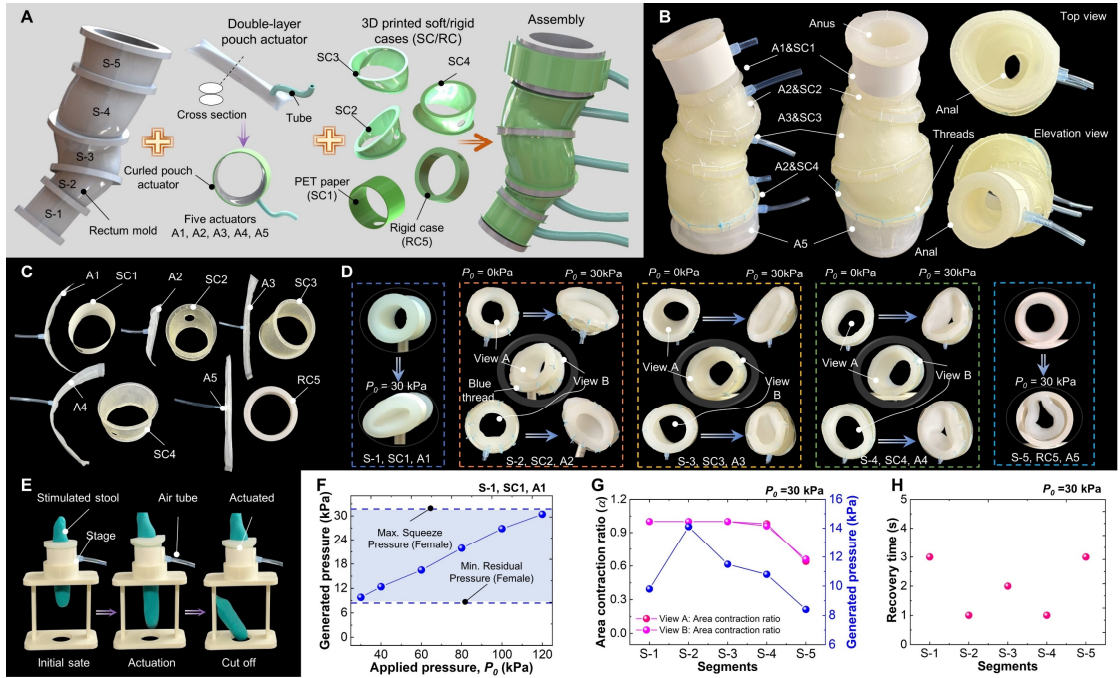

Fig. 5 Rectal module: rectum mold and its segments, soft cases, rigid case, and actuators. **A:** Components and assembly process for the rectal module. **B:** Assembled prototype from top and elevation views. **C:** Optical images of five pouch actuators (A1~A5), soft cases (SC1~SC4), and hard cover (RC5). **D:** Photos of deformation of the pouch actuators between their initial and activated states. **E:** Demonstration of the small system (S-1, SC1, and A) severing simulated fecal matter. **F:** Comparison of pressures generated by the system (S-1, SC1, and A1) against the maximum squeeze pressure and minimum residual pressure typically observed in the female anal region. **G:** Performance comparison of each segment inflated by actuators in terms of contraction ratios. **H:** Recovery time required for the actuators to return to their original shape.

## 2.5 Defecation robotic system

We designed the defecatory system, power supply system, pressure sensing system and data acquisition system, respectively (Fig. 6). The overall system includes mechatronics, flush, stage, and rectal module (Fig. 6**A**). Fig. 6**B** shows the schematics of power supply, data acquisition system and control systems. The pneumatic supply system comprises an air compressor (SR-L30MPT-01), a combined pressure controller and filter unit (FRF300-03-MD), electromagnetic valves (CKD Company, MEVT series), a Koganei solenoid valve (030E1-L DC24V), an RS PRO switching power supply (PDF-600-24), a controller (Arduino Mega), pressure sensors (PSE564-A2-28), and a data acquisition system (LABVIEW, NI instruments). Throughout the experimental phase, the input pressure was meticulously adjusted to 500 kPa. We have innovatively designed five pressure-regulating valves and solenoid valves to ensure precise gas supply and exhaust for each soft actuator. For accurate measurement of the input pressure, five pressure sensors were utilized. Data acquisition is conducted via the LABVIEW DAQ system for pressure metrics, while the simulated weight is quantified using a precision balance (FigsS.3).

In this study, A1 and A5 were primarily employed for the opening and closing of the anus, as well as for preventing the retrograde movement of simulated faeces. A2, A3, and A4 were utilized to generate peristaltic motion for faeces transportation. We conducted tests on the movement capabilities of each simulator component (Fig. 6**C**). It can be observed that upon activation of the various actuators, the interior of the rectum model underwent corresponding deformations. A1 was able to effectively seal the anus with an input pressure of 30 kPa. During actual operation, control modes for A1 and A5 were configured to operate independently, whereas A2, A3, and A4 were managed through a pattern of inflation and deflation at predetermined intervals. The trajectory of defecating the cylindrical faeces can be observed from the top and front views (Fig. 6**D**).

We formulated simulated faeces with a specified viscosity by combining clay and olive oil, subsequently evaluating their defecation performance. Fig. 6**E** shows the robot achieves significant defecation capabilities when the mass ratio reaches 9%, with the utilized faeces being cylindrical, having a diameter of 20 mm and a length of 70 mm (Video S5). We selected a 9% ratio for further investigation. Additionally, we examined the influence of applied pressure ($p$) and one-quarter of the period ($t$) on the defecation efficacy of our robotic system (Fig. 6**F** & **G**). It can be concluded that the defecation speed consistently increases with the escalation of applied pressure when $t$ is maintained at 1 s. Moreover, we investigated the impact of $t$ on the defecation performance. The defecation speed noticeably declines, indicating that faeces are more effectively expelled along the inner wall of the rectum as the actuator's squeezing speed increases. Consequently, the optimal values for pressure and t, determined to be 30 kPa and 1 s respectively, are chosen to maximize defecation velocity, closely mirroring the frequency of human defecation as evidenced by magnetic resonance imaging (MRI) findings. We also prepared faeces samples of various diameters, lengths, and shapes (FigS. 4). Fig. 6**H** shows that both spherical and cylindrical shapes (diameter < 30 mm) can be expelled by our robot, whereas larger specimens (diameter > 30 mm) could not be expelled, attributed to the anus diameter being 30 mm.

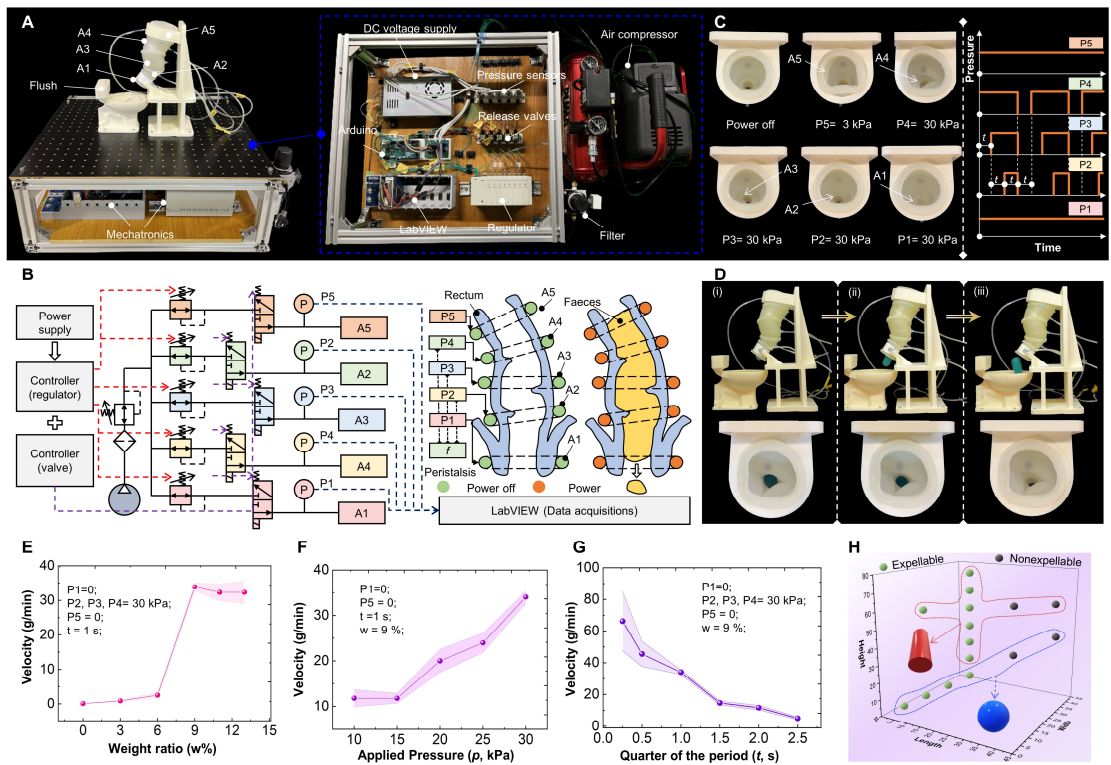

Fig. 6 Defecation robotic system and associated components. **A:** Optical images of the defecation robot along with its pneumatic systems. **B:** Detailed schematics highlight the power supply, data acquisition system, and control system. **C:** Top view revealing the movements of each actuator in relation to the input pressure signals designated for each actuator. **D:** Optical photos of the defecation process from both top and front perspectives. **E:** Relationship between defecation speed and the weight ratio of oil to clay. **F:** Correlation between defecation speed and applied pressure. **G:** Relationship between defecation speed and control time (*t*: quarter of the period). **H:** Defecation performance of different types of faeces with various shapes.

When humans encounter FI, various situations may arise, such as diarrhea, constipation resulting the anal to cut stool, and occasional passage of excessively long stools. To address these issues, we have employed our robots to simulate and replicate these scenarios. For the diarrhea (Fig. 7**A**& Video S6), we synthesized liquid feces using sodium alginate and water. In this scenario, actuators A2, A3, A4, and A5 were rendered non-operational, while A1 was closed for a period before being opened to simulate this condition. Fig. 7**B** illustrates the effective sealing capability of our A1 actuator, preventing any leakage. The liquid feces can be rapidly expelled into the flush upon the activation of the A1 actuator. Experiments with pure water were also conducted (FigS.5). Moreover, our robot can simulate the human anal canal's ability to cut feces (Fig. 7**C**). Here, we adjusted A5 to 1 kPa and set A4, A3, and A2 to simulate normal peristaltic motion at 30 kPa, with a control time (*t*) of 1 second each. After several peristalsis cycles, A1 initiated the feces cutting process. The length and type of feces were regulated by the PWM wave duration applied to A1. Short feces were produced when $t_0$, $t_1$ and $t_2$ were set to 12 seconds, 2 seconds (on), and 2 seconds (off), respectively (Fig. 7**D**). Conversely, extremely long feces were achieved by setting $t_0$, $t_1$ and $t_2$ to 0 seconds, 1 second (on), and 1 second (off), respectively (Fig. 7**E&F**). In this way, A1, A2, A3 and A4 work coordinately for the peristaltic motion. Also, our robot can realize multiple faeces cutting actions by mimicking the human anal (FigS.5). It can be found that our robot is flexible, suitable for both liquid and solid faeces, capable of discharging extremely long faeces, and incorporates the peristaltic characteristics of the rectum. It has advantages of low entirely soft, input pressure, long lifespan, and fast defecation speed. Finally, we compared our defecation robot or simulator with others and found that our robot is flexible, suitable for both liquid and solid feces, capable of discharging extremely long feces, and incorporates the peristaltic characteristics of the rectum (Table S2).

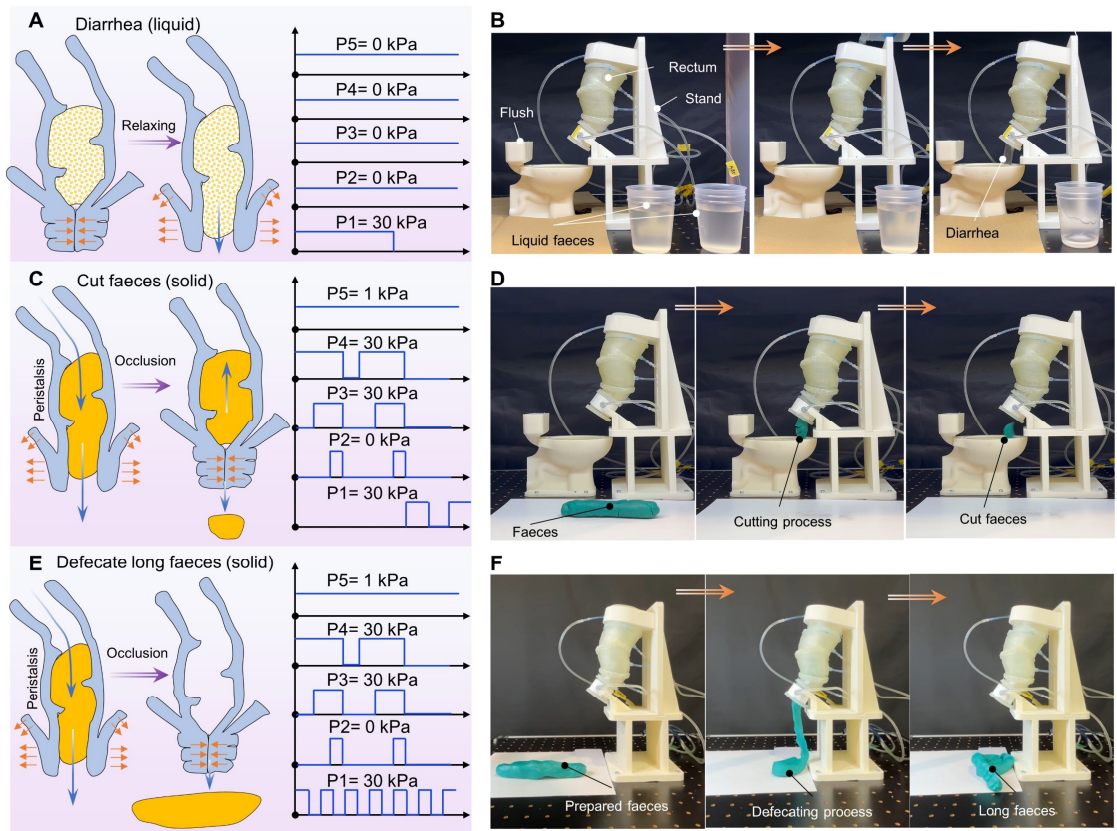

Fig. 7 Simulating three unique defecation scenarios encountered in humans using our developed defection robot. **A:** Emulation of human diarrhea and the operational control strategies for each actuator. **B:** Optical images showcasing our defecation robot replicating the condition of human diarrhea. **C:** Motion control in the simulator's anal mechanism for cutting solid feces. **D:** Optical images showing our defecation robot replicating severing solid feces. **E:** Replication of scenarios involving the expulsion of extremely long solid feces, accompanied by detailed control strategies. **F:** Optical images showing our defecation robot enables to severe extremely long faeces.

## 3. Conclusion

In summary, this study introduces a rectum-inspired, machine-learning-enhanced soft robotic system aimed at replicating three distinct defecation scenarios to investigate fecal incontinence. The system encompasses a power supply, pressure sensing, and data acquisition systems, along with a flushing mechanism, stage, and a rectal module designed to mimic the peristaltic movement of fecal matter. The rectal module integrates a rectum model with actuators inspired by AAS muscles and is encased in both rigid and soft materials. Our evaluation of various soft materials, through stress-strain analysis, identified Ecoflex 00-30 as most closely mirroring the stress and strain properties of the human rectum. The replication mold method facilitated the fabrication of both the rectum model and its segments. Additionally, we developed, produced, and assessed the AAS muscle-inspired actuators, optimizing their functionality based on six criteria: pressure, structure, covering material, length, width, and turns. Initial modeling was achieved using a multilayer perceptron (MLP) model, which provided highly accurate predictions ($R^2$ near 1), enabling the identification of optimal configurations within physical limitations.

Subsequently, the optimal parameters for the double-layer pouch actuator (Type 4) sheathed in PET paper were integrated into the rectal module, significantly enhancing contraction ratios, pressure generation, and reducing recovery times. The completed robotic system effectively simulates peristaltic movements within the rectum and the occlusion of the anal canal, replicating common human conditions such as diarrhea, normal feces, and scenarios of prolonged fecal discharge under varied control strategies. This innovative robot paves the way for new research into constipation-related factors through physical simulation and aids in the development of defecation-assistive devices for the elderly.

## Acknowledgments
We thank the helpful discussion from Prof. Young Ah eong, Prof. Toshinori Fujie, Prof. Kenjiro Tadakuma, Prof. Hideyuki Sawada.

## Funding
This work was supported by Grant-in-Aid for Scientific Research on Innovative Areas (Research in a proposed research area) from the Japan Society for the Promotion of Science (Grant Nos. 18H05473 and 23K13290).

## Author Contributions
Z.M., S.S., and H.N. were responsible for the conceptualization, design, fabrication, and performance evaluation of the devices, in addition to writing and managing the project. S.M., K.S., and S.M. supervised the projects and facilitated the acquisition of funding. All authors have read and approved the final manuscript.

## Conflicts of Interest
The authors declare that there are no competing interests.

## Data Availability
Data supporting the findings of this study are available in the main text or the Supplementary Materials

## Supplementary Materials
FigS. 1: Molds, fabrication methods and equipment in this study; Stress-strain characteristics of the printed material; photos of AR-M2 samples.
FigS. 2 Manufacturing processes of four type actuators.
FigS. 3: LabVIEW interface.
FigS. 4 Design and fabrication of stimulated faeces.
FigS. 5 Water defecation and cutting faeces twice
TableS.1: Performance comparison of actuators
TableS.2: Performance comparison of defecation robots/simulator
Video S0: code execution
Video S1: Four types of pouch actuators
Video S2: Fabrication process
Video S3: Rectum module
Video S4: Demo to cut off stimulated faeces
Video S5: Defecation of the simulated faeces with/without lubricant
Video S6: Simulating three unique defecation scenarios encountered in humans